\documentclass[letterpaper]{article} % DO NOT CHANGE THIS
\usepackage{aaai23}  % DO NOT CHANGE THIS
\usepackage{times}  % DO NOT CHANGE THIS
\usepackage{helvet}  % DO NOT CHANGE THIS
\usepackage{courier}  % DO NOT CHANGE THIS
\usepackage[hyphens]{url}  % DO NOT CHANGE THIS
\usepackage{graphicx} % DO NOT CHANGE THIS
\usepackage{natbib}  % DO NOT CHANGE THIS AND DO NOT ADD ANY OPTIONS TO IT
\usepackage{caption} % DO NOT CHANGE THIS AND DO NOT ADD ANY OPTIONS TO IT
\usepackage{algorithm}
\usepackage{algorithmic}

\usepackage{newfloat}
\usepackage{listings}
\DeclareCaptionStyle{ruled}{labelfont=normalfont,labelsep=colon,strut=off} % DO NOT CHANGE THIS
\floatstyle{ruled}
\newfloat{listing}{tb}{lst}{}
\floatname{listing}{Listing}
\title{LiT Tuned Models for Efficient Species Detection}
\author {
    Andre Nakkab, %\textsuperscript{\rm 1}
    Benjamin Feuer, % \textsuperscript{\rm 1}
    Chinmay Hegde %\textsuperscript{\rm 1}
}
\affiliations {
    New York University Tandon School of Engineering\\
    ajn313@nyu.edu, bf996@nyu.edu, chinmay.h@nyu.edu
}

\begin{document}

\maketitle

\begin{abstract}
\begin{quote}
Recent advances in training vision-language models have demonstrated unprecedented robustness and transfer learning effectiveness; however, standard computer vision datasets are image-only, and therefore not well adapted to such training methods. Our paper introduces a simple methodology for adapting any fine-grained image classification dataset for distributed vision-language pretraining. We implement this methodology on the challenging iNaturalist-2021 dataset, comprised of approximately 2.7 million images of macro-organisms across 10,000 classes, and achieve a new state-of-the art model in terms of zero-shot classification accuracy. Somewhat surprisingly, our model (trained using a new method called locked-image text tuning) uses a pre-trained, frozen vision representation,  proving that language alignment alone can attain strong transfer learning performance, even on fractious, long-tailed datasets. Our approach opens the door for utilizing high quality vision-language pretrained models in agriculturally relevant applications involving species detection.
\end{quote}
\end{abstract}

\section{Introduction}
    Deep neural network models that have been trained on both vision and text data have enjoyed considerable success in various applications. These so-called `vision-language' models perform exceedingly well on general computer vision tasks (such as classification on the ImageNet dataset). However, such models are extremely resource-intensive, and training them from scratch for each new dataset can be cumbersome. 
    
    How, then, can we reduce these costs? One approach is via the well-established route of \emph{transfer learning}. By selecting pre-trained models and fine-tuning them to more specialized tasks, we can achieve competitive accuracy for a variety of tasks. However, in the case of vision-language models, this itself poses a challenge; several standard computer vision datasets are image-only (without natural-language caption information). For many agricultural applications where text descriptions are absent, this limitation can prove crucial.

\section{Our Contributions}
    
    We introduce a simple methodology for adapting existing image-only datasets for vision-language (VL) pretraining. The high level idea is to automatically generate per-image captions using tagged metadata (including class names, category names, and taxonomy information). Using this language-augmented dataset, we can fine-tune existing VL models. 
    
    We validate this methodology on the iNaturalist 2021 (iNat2021) dataset. The task is macro-organism classification (or species identification), and therefore is relevant to field biologists or other agricultural domain experts. However, images in the iNat2021 dataset have been collected via `citizen scientists' in a highly unstructured manner, and descriptive natural language captions for each image are not available. To remedy this, we use our automatic caption generation procedure to create a new dataset, which we call iNat-Captions. We use this dataset to obtain new LiT-tuned models that achieve a new state-of-the-art for zero-shot classification accuracy on iNat2021 using VL-training; see Table~\ref{tab:results}. 

    Somewhat surprisingly, we find that it is not necessary to adjust the weights of the vision tower. Using a technique called Locked-image Text (LiT) tuning, we show that aligning the language representation to the new captions alone suffices for competitive zero-shot classification performance, therefore cutting training costs significantly. To the best of our knowledge, our results are also the first to demonstrate the power of LiT tuning on fractious, long-tailed datasets such as iNaturalist.

\section{Background}	
	Since the recent introduction of CLIP, there has been a push toward developing open-source frameworks for training vision-language models (Wortsman et al. 2022). By training both an image classification model and a natural language processing model in tandem, these architectures are able to learn image classification tasks without the use of manually applied ground truth labels, instead using natural language captions or tags (Radford et al. 2021). Thus, these architectures can be considered semi-supervised, meaning that there is considerably less work/cost involved in generating usable datasets for training, since there is little to no manual labeling of data involved. 
 
 Recent empirical studies have demonstrated that contrastive vision-language models are considerably more robust to label noise than classifiers trained with softmax + cross-entropy loss (Feuer et al. 2022). Additional experimental studies have demonstrated that it was possible to use a pretrained, locked image tower with far fewer samples and achieve results close to, but not on par with, vision-language models trained on semi-supervised data. (Zhai et al. 2022)
 
 	iNaturalist 2021 is comprised of approximately 2.7 million images across 10,000 categories. Each category represents a single species. All kingdoms of macro-organisms are represented (i.e., mammals, fish, fungi, birds, trees, etc.) This poses an intriguing challenge for an image classification model, as individual images may involve elements from multiple classes, e.g., an insect standing on a plant, or a bird in a tree. Beyond the intrinsic difficulty of a 10,000-class categorization task, macro-organism identification represents further challenges.
  
\section{Dataset Creation}
	We start with the iNaturalist Species Classification and Detection Dataset for 2021 (Van Horn et al. 2018).   Since iNaturalist is not a captioned dataset, we devise a method for adapting the dataset to LiT-Tuning. The optimal method for automatically captioning a dataset for which no captions exist is an open problem. In the original CLIP paper, Radford et al use an ensemble of 80 captions with the classname (and sometimes the object supercategory) varying each time; for example, ``A photo of a clam, a type of seafood". Experimental results indicate that while this method is suitable for inference, it is not optimal for pretraining, owing to the homogeneity of the caption space. Furthermore, ``bag of words" VL models are largely agnostic to word order and sentence syntax (Feuer et al. 2022). After some testing, we selected a simple method which can be deployed on any dataset for which class-level metadata exists:
    \begin{enumerate}
    \item Aggregate metadata for each class.
    \item Select a subset of metadata columns which maximize inter-class differences
    \item Generate image captions as `A photo of a $\langle$ CONCAT\_METADATA $\rangle$'
    \end{enumerate}
 
 The iNat2021 dataset provides a rich range of metadata including common species name, general description, and binomial scientific classification of the organism in the image. The resulting dataset is stored in webdataset format for portability across multiple platforms. The dataset is publicly available in a repository on Hugging Face linked below.
 
	\begin{figure}
        \centering
        \begin{tabular}{ccc}
        \includegraphics[width=2.3cm]{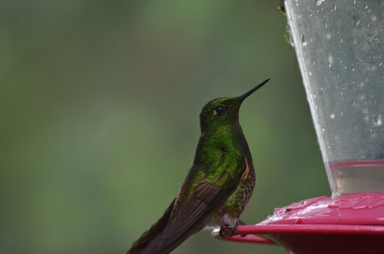} & \includegraphics[width=2.3cm]{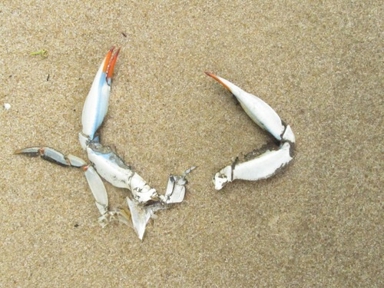} & \includegraphics[width=2.3cm]{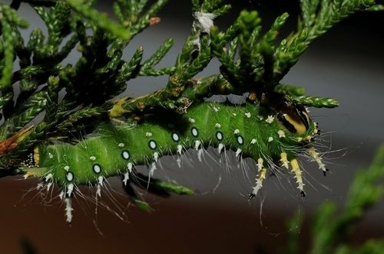} \\
        \includegraphics[width=2.3cm]{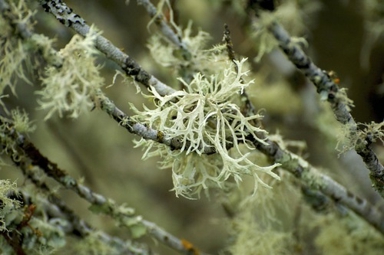} & \includegraphics[width=2.3cm]{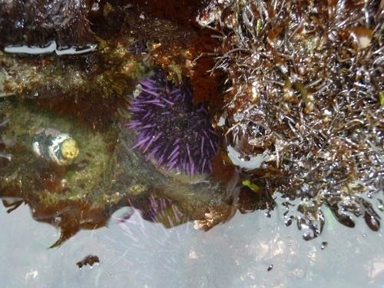} & \includegraphics[width=2.3cm]{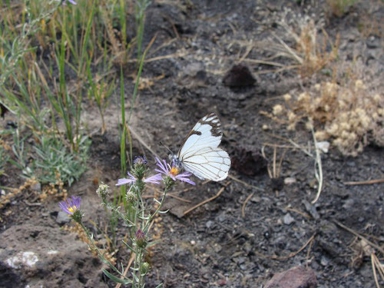}
        \end{tabular}
        \caption{Samples from the iNaturalist 2021 dataset, respectively captioned: \emph{``A photo of the Buff-tailed Coronet Birds Boissonneaua flavescens"}; \emph{``A photo of the Common Blue Crab Animalia Callinectes sapidus"}; \emph{``A photo of the Imperial Moth Insects Eacles imperialis"}; \emph{``A photo of the Oakmoss Fungi Evernia prunastri"}; \emph{``A photo of the Pacific Purple Sea Urchin Animalia Strongylocentrotus purpuratus"}; \emph{``A photo of the Pine White Insects Neophasia menapia"}}
    \end{figure}
    %\begin{figure}
    %    \centering
    %    \includegraphics[width=8cm]{graphs/00518418.jpg}
    %    \caption{Sample from the iNaturalist 2021 dataset, captioned: \emph{``A photo of the arrowleaf ragwort Plants Senecio triangularis"}}
    %\end{figure}

	%\emph{Data will be made available (?) @ChinmayHegde will have to answer this}
\section{Training}
    The models in this paper were LiT-tuned using VLHub, a fork of OpenCLIP with added support for additional architectures and pretraining techniques (Feuer et al. 2022).  Primary comparative ablations across models included the learning rate warm-up and the model used as the image tower. 
    We chose LiT-tuning because the technique requires fewer samples and time than pretraining from scratch, making it more suitable for resource-constrained agricultural contexts.
    
    The image tower used in the learning rate warm-up experiments was a large-scale Vision Transformer (ViT-Large) with approximately 307 million trainable parameters. This model provides a vision backbone which is comparable to state of the art convolutional neural networks, while being considerably more computationally efficient (Dosovitskiy et al. 2021).
    The language tower used was a text encoder with approximately 33 million trainable parameters, capable of taking natural language text inputs, as well as user-defined tags.
    
    Training portability was improved through the use of a Singularity container equipped with the appropriate Miniconda environment. Additionally, a launcher script was created to facilitate the use of the container in a distributed training environment using SLURM. Models were trained using resources from the National Science Foundation's XSEDE project prior to its transition into ACCESS. Training hardware included 16 NVIDIA V100 GPUs distributed across 2 nodes. Code base can be found in a GitHub repository linked below.

\section{Experimental Results}
    Experiments were performed for comparative fine-tuning of hyperparameters, primarily to test the hypothesis that scaling the learning rate warm-up period linearly would improve performance, but overshooting would lose many of the benefits of warm-up. The iNaturalist 2021 validation data set was then used to test zero-shot transfer performance.
    
    An initial model utilizing a ViT-Large image backbone and a warm-up period of 2000 was the first experiment attempted. This model achieved a top-1 accuracy of 61.96\% and a top-5 accuracy of 86.71\% after 18 epochs. 
    
    The second model was trained using a learning rate warm-up period of 1000, i.e., the maximum learning rate was reached more quickly. This model was able to converge more quickly, and provided slightly higher performance, reaching a top-1 accuracy of 63.28\% and a top-5 accuracy of 87.48\% after 18 epochs.
    
    Finally, an experiment was run with a warm-up period of 500, which led to the network being unable to successfully converge during training. This is likely due to the model taking large, early gradient steps in an inappropriate direction, thereby becoming unable to continue optimization in the time allotted. This can be largely avoided using an appropriate warm-up period, which is in keeping with the original hypothesis. 
    
  Using this approach, we already achieve competitive performance with previously existing methods. Our method far out-performs the top performing reported methods on the Semi-iNat 2021 dataset, which is easier in the sense that it includes fewer classes, but harder in the sense that only coarse labels and metadata are provided. On the other end of the spectrum, we have fully supervised models, which are capable of reaching greater than 95\% accuracy on iNaturalist 2021, but do so by training the image tower from scratch, which requires more labeled samples and greater computational resources than our method. Comparative results can be found in Table 1. 
    
    In all, we believe that LiT-tuning is the ideal approach when large amounts of coarsely labeled data and small amounts of finely labeled data are available, a context which occurs often in practice.
    \begin{figure}
        \centering
        \includegraphics[width=6cm]{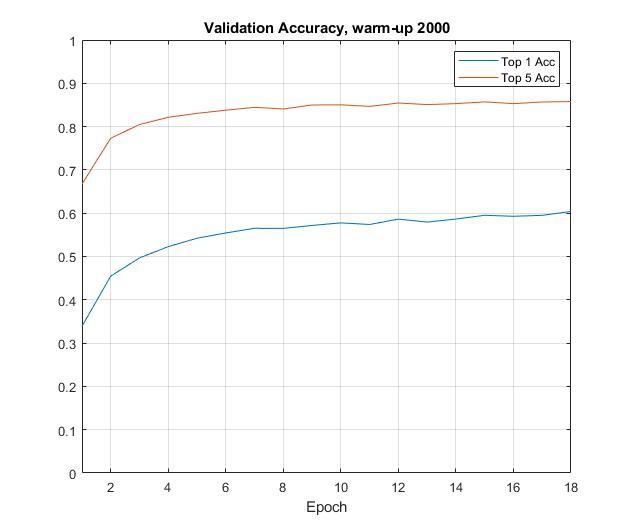}
        \caption{Accuracy per epoch for first model}
    \end{figure}
    \begin{figure}
        \centering
        \includegraphics[width=6cm]{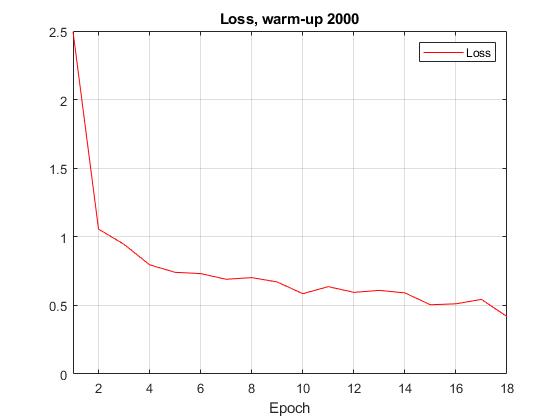}
        \caption{Loss per epoch for first model}
    \end{figure}
    \begin{figure}
        \centering
        \includegraphics[width=6cm]{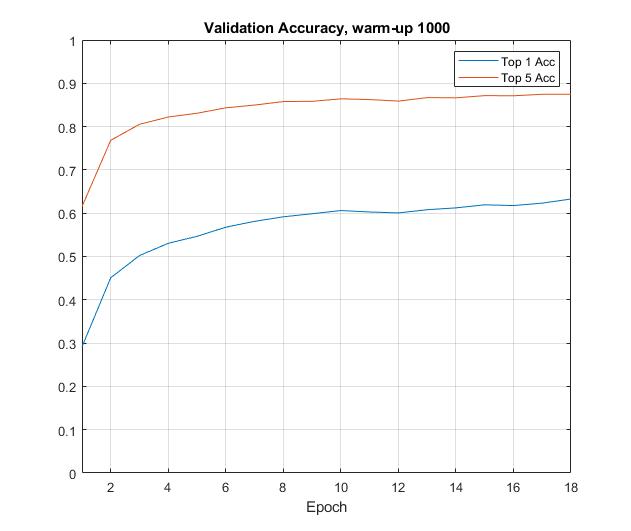}
        \caption{Accuracy per epoch for second model}
    \end{figure}
    \begin{figure}
        \centering
        \includegraphics[width=6cm]{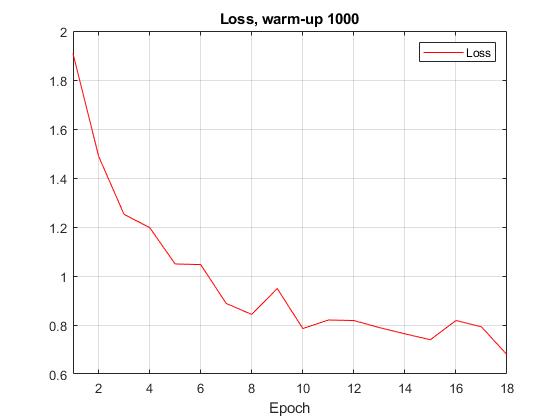}
        \caption{Loss per epoch for second model}
    \end{figure}
    
    \begin{table*}
    \centering
\begin{tabular}{||c | c | c | c||} 
 \hline
 Method & Image Tower & Top-1 accuracy (\%) & Top-5 accuracy (\%) \\ [0.5ex] 
 \hline
 Full Training & ResNet18 & 26.35 & -  \\
 \hline
 Full Training & ResNet34 & 31.02 & - \\
 \hline
 2D-DWT Transfer & WaveMix-Lite-256/7 & 33.23 & - \\ 
 \hline
 CoMatch+CCSSL & ResNet50 & 39.85 & 63.68 \\
 \hline
 FixMatch & ResNet50 & 47.9 & -\\
 \hline
 Full Training & ResNet50 & 61.6 & 81.8\\
 \hline
 ImageNet Pre-training & ResNet50 & \textbf{65.4} & 85.1 \\
 \hline
  \textbf{LiT (ours)} & \textbf{OpenCLIP (ViT)} & \textbf{63.28} & \textbf{87.48}\\
 \hline
\end{tabular}
\caption{Comparative results on iNat 2021 / Semi-iNat. Our model achieves a new state-of-the-art for VL zero-shot classification on iNat2021. Performance is comparable to a ResNet50 model pre-trained on ImageNet and fine-tuned on iNat.}
\label{tab:results}
\end{table*}
%\end{center}
    
\section{Discussion}
	The effects seen by shortening the learning rate warm-up period were in line with the original hypothesis. This is likely explained by a mechanism similar to learning rate scaling with batch size. As previously reported in the literature, increasing the learning rate mitigates the effect of larger batch sizes, but leading to lower accuracy by increasing the size of the gradient and decreasing learning effects from individual samples of input data (Goyal et al. 2017).
 
	% We can observe this mathematically by considering the mini-batch SGD update rule. Let \(B_i\) represent a minibatch of size \(j\), \(w_t\) represent the current weight, and \(\hat{\eta}\) represent the effective learning rate: \\
	% \(w_{t+1}=w_{t}-\hat{\eta}\frac{1}{j}\sum_{i<j} \nabla L(B_i,w_t)
	% \)\\
 %    It is intuitive, then, that by scaling the learning rate linearly with batch size such that \(\hat{\eta}=\eta*j\), we can mitigate the effects of a larger batch size by cancellation, i.e.:\\
 %    \(w_{t+1}=w_{t}-\eta*j\frac{1}{j}\sum_{i<j} \nabla L(B_i,w_t)\\=w_{t}-\eta\sum_{i<j} \nabla L(B_i,w_t)
	% \)\\
 %    This is likely not an exact method due to stochasticity, but is highly effective in most scenarios. 
    
    Therefore, by reducing the warm-up period as compared to prior implementations of LiT-tuning, we are able to reach our maximum learning rate more quickly, scaling it more closely with batch size. This still prevents the model from getting ``lost" in the gradient by starting with a smaller learning rate, but will likely lead to faster convergence, thereby improving training efficiency without noticeably sacrificing performance.
    
    However, it should be noted that too large a scaling factor will mitigate these benefits, as even the learning rates located earlier in the warm-up period will be increased. Our experimentation reveals that this can often lead to models which do not converge.
    
  Our most robust model was achieved by combining these techniques, i.e., starting with a small learning rate, and increasing by epoch until we reach an appropriately scaled learning rate for the given batch size. This is in line with the findings of earlier studies involving other models, but gives important confirmation that this technique is also effective when applied to LiT-tuning.

\section{Conclusions and Future Work}

   We develop a simple approach for training VL models on image-only datasets, and validate it for the iNaturalist 2021 dataset; this is important for agriculurally relevant applications such as species detection.
   
   Further experimental exploration should be done comparing the performance of different image towers on the performance of the model overall. The first priority would likely be the Swin transformer, which improves on the techniques seen in ViT, including the use of a local attention mechanism rather than a global attention mechanism (Liu et al. 2021). 
    
    Additionally, more detailed exploration of the inference capabilities of the models is a natural next step. This may include determining which classes/species (or genera, families, orders, etc.) prove most difficult for each model to identify. This would provide insight into the strengths of certain design elements over others. 
    
    Additionally, a portable implementation for use in actual field testing would likely be necessary for wide deployment. Embedding the highest performing models into a mobile or web application could provide information on which model is most robust when released into the wild.
    
\section{References:}
\begin{enumerate}
    \item[] Radford, A., Kim, J. W., Hallacy, C., Ramesh, A., Goh, G., Agarwal, S., ... and Sutskever, I. (2021). ``Learning transferable visual models from natural language supervision." In \emph{International Conference on Machine Learning} (pp. 8748-8763). PMLR.
    \item[]  Pan, S. J., Yang, Q. (2010). ``A Survey on Transfer Learning." In \emph{IEEE Transactions on Knowledge and Data Engineering}, vol. 22, no. 10, pp. 1345-1359, Oct. 2010, doi: 10.1109/TKDE.2009.191.
    \item[] Zhai, X., Wang, X., Mustafa, B., Steiner, A., Keysers, D., Kolesnikov, A., and Beyer, L. (2022). ``Lit: Zero-shot transfer with locked-image text tuning." In \emph{Proceedings of the IEEE/CVF Conference on Computer Vision and Pattern Recognition} (pp. 18123-18133).
    \item[] Van Horn, G., Mac Aodha, O., Song, Y., Cui, Y., Sun, C., Shepard, A., ... and Belongie, S. (2018). ``The iNaturalist species classification and detection dataset." In \emph{Proceedings of the IEEE conference on computer vision and pattern recognition} (pp. 8769-8778).
    \item[] Goyal, P., Dollár, P., Girshick, R., Noordhuis, P., Wesolowski, L., Kyrola, A., and He, K. (2017). ``Accurate, large minibatch SGD: Training imagenet in 1 hour." arXiv preprint arXiv:1706.02677.
    \item[] Dosovitskiy, A., Beyer, L., Kolesnikov, A., Weissenborn, D., Zhai, X., Unterthiner, T., Dehghani, M., Minderer, M., Heigold, G., Gelly, S., Uszkoreit, J., and Houlsby, N. (2021). ``An Image is Worth 16x16 Words: Transformers for Image Recognition at Scale." ArXiv, abs/2010.11929.
    \item[] Liu, Z., Lin, Y., Cao, Y., Hu, H., Wei, Y., Zhang, Z., ... and Guo, B. (2021). ``Swin transformer: Hierarchical vision transformer using shifted windows." In \emph{Proceedings of the IEEE/CVF International Conference on Computer Vision} (pp. 10012-10022).
    \item[] Wortsman, M., Ilharco, G., Kim, J. W., Li, M., Kornblith, S., Roelofs, R., ... and Schmidt, L. (2022). Robust fine-tuning of zero-shot models. In \emph{Proceedings of the IEEE/CVF Conference on Computer Vision and Pattern Recognition} (pp. 7959-7971).
    \item[] Feuer, B., Joshi, A., and Hegde, C. (2022). Caption supervision enables robust learners. \emph{arXiv preprint arXiv}:2210.07396.
    \item[] Su, J. C., and Maji, S. (2021). The semi-supervised inaturalist challenge at the FGVC8 workshop. \emph{arXiv preprint arXiv}:2106.01364.
    \item[] Jeevan, P., Viswanathan, K., and Sethi, A. (2022). WaveMix-Lite: A Resource-efficient Neural Network for Image Analysis. \emph{arXiv preprint arXiv}:2205.14375.
    \item[] Van Horn, G., Cole, E., Beery, S., Wilber, K., Belongie, S., and Mac Aodha, O. (2021). Benchmarking representation learning for natural world image collections. In \emph{Proceedings of the IEEE/CVF conference on computer vision and pattern recognition} (pp. 12884-12893).
    \item[] Yang, F., Wu, K., Zhang, S., Jiang, G., Liu, Y., Zheng, F., ... and Zeng, L. (2022). Class-Aware Contrastive Semi-Supervised Learning. In \emph{Proceedings of the IEEE/CVF Conference on Computer Vision and Pattern Recognition} (pp. 14421-14430).
    \item[] Su, J. C., and Maji, S. (2021). Semi-Supervised Learning with Taxonomic Labels. \emph{arXiv preprint arXiv}:2111.11595.
    %\item[] NYU OpenCLIP: https://github.com/NYU-DICE-Lab/open\_clip
    \\
    \item[] Dataset Repository: \\ https://huggingface.co/ajn313/inat\_captions/tree/main 
    \\
    \item[] Code Base Repository: \\ https://github.com/NYU-DICE-Lab/open\_clip
    
\end{enumerate}

\end{document}